\documentclass{article}
\usepackage{ijcai20}

\usepackage{times}
\usepackage{soul}
\usepackage{url}
\usepackage[hidelinks]{hyperref}
\usepackage[utf8]{inputenc}
\usepackage[small]{caption}
\usepackage{graphicx}
\usepackage{amsmath}
\usepackage{amsthm}
\usepackage{booktabs}
\usepackage{algorithm}
\usepackage[noend]{algorithmic}
\usepackage{hyperref}
\usepackage{url}
\usepackage{graphicx} % DO NOT CHANGE THIS

\usepackage{amsfonts}

\usepackage{amsmath}
\usepackage{wrapfig}

\usepackage{soul}

\usepackage{pdfpages}

\newcommand{\squishlist}{
 \begin{list}{$\bullet$}
  { \setlength{\itemsep}{0pt}
     \setlength{\parsep}{2pt}
     \setlength{\topsep}{2pt}
     \setlength{\partopsep}{0pt}
     \setlength{\leftmargin}{1em}
     \setlength{\labelwidth}{1em}
     \setlength{\labelsep}{0.5em} } }

     \newcommand{\squishend}{
  \end{list}  }

\newcommand{\changespace}[1]{\renewcommand{\baselinestretch}
	{#1}\normalsize}
\changespace{0.96}

\usepackage{titlesec}

\makeatletter
\@addtoreset{section}{part}
\makeatother
\titleformat{\part}[display]
{\normalfont\LARGE\bfseries\centering}{}{0pt}{}

\author{
	Rajiv Ranjan Kumar\footnote{Contact Author}\and
	Pradeep Varakantham\\
	\affiliations
	School of Information Systems, 
	Singapore Management University\\
	\emails
	\{rajivrk.2017@phdcs.smu.edu.sg, pradeepv@smu.edu.sg\}
}

\begin{document}
	\title{On Solving Cooperative MARL Problems with a Few Good Experiences}
	\maketitle
	
	\begin{abstract}
		Cooperative Multi-agent Reinforcement Learning (MARL) is crucial for cooperative decentralized decision learning in many domains such as search and rescue, drone surveillance, package delivery and fire fighting problems. In these  domains, a key challenge is  learning with a few good experiences, i.e., positive reinforcements are obtained only in a few situations (e.g., on extinguishing a fire or tracking a crime or delivering a package) and in most other situations there is zero or negative reinforcement. Learning decisions with a few good experiences is extremely challenging in cooperative MARL problems due to three reasons. First, compared to the single agent case, exploration is harder as multiple agents have to be coordinated to receive a good experience. Second, environment is not stationary as all the agents are learning at the same time (and hence change policies). Third, scale of problem increases significantly with every additional agent.  
		
		Relevant existing work is extensive and has focussed on dealing with a few good experiences in single-agent RL problems or on scalable approaches for handling non-stationarity in MARL problems. Unfortunately, neither of these approaches (or their extensions) are able to address the problem of sparse good experiences effectively. Therefore, we provide a novel fictitious self imitation approach that is able to simultaneously handle non-stationarity and sparse good experiences in a scalable manner. Finally, we provide a thorough comparison (experimental or descriptive) against relevant cooperative MARL algorithms to demonstrate the utility of our approach.
	\end{abstract}
	
	\section{Introduction}
	Cooperative MARL is an important framework for learning agent policies in multiple domains such as disaster rescue~\cite{parker2016exploiting}, fire fighting~\cite{oliehoek2008optimal} and package delivery (box pushing)~\cite{seuken2012improved}. In these problems, a team of decentralized agents coordinate to accomplish tasks (find people, extinguish fires, and deliver boxes to destinations) in uncertain domains. There are multiple key challenges in these learning problems: (a) Uncertainty in movement or in accomplishing tasks; (b) Coordination of decentralized entities to accomplish tasks (e.g., big fires require multiple fire engines or delivering a large box may require multiple robots; (c) Affected global state: Global state (representing status of tasks) can be impacted by agent actions; and most importantly (d) Sparse good experiences: rewards are obtained only when tasks are accomplished and there are only a few tasks. 
	
	The problem of learning with a few good experiences or sparse rewards studied also in single agent RL~\cite{oh2018self} is exacerbated in MARL problems due to three reasons: (1)  Exploration is significantly harder as multiple agents have to be coordinated;  (2) Environment is not stationary (multiple agents are learning together); and (3) Scale of problem increases significantly with every additional agent. In summary, \textbf{\em exploration to find good policies is challenging and even if we find good policies, addressing non-stationarity and  scalability can result in forgetting those good policies.} 
	
	Research of relevance to this paper has focussed on addressing: (a) A few good experiences~\cite{pomerleau1991efficient,oh2018self,lee2019improved,lerer2019learning} primarily in single agent RL and sparsely in multi-agent RL through imitation learning; (b) Non-stationarity (due to multiple agents learning simultaneously)  in MARL~\cite{palmer2018lenient,omidshafiei2017deep,foerster2018counterfactual}; (c) Scalability in MARL by exploiting anonymity and homogeneity~\cite{nguyen2017policy,yang2018deep}. Even though the relevant research in MARL is extensive, there is not much research on handling sparse good experiences in MARL. Most importantly, the current best approaches  are unable to provide good policies (as demonstrated in experimental results) for cooperative MARL problems with only a few good experiences. 
	
	To that end, we provide a novel approach that not only learns effectively from a few good experiences but is also decentralized and scalable.  Specifically, we make the following key contributions: (i) we incorporate self imitation into a state of the art MARL approach called Neural Fictitious Self Play (NFSP), so as to replay past good experiences and ensure effective exploration;  (ii) we introduce a modification to policy averaging in NFSP to ensure good policies remain relevant; (iii) we also provide theoretical intuition for why the new policy averaging method follows the generalized weakened fictitious play property, thereby guaranteeing convergence. Finally, we demonstrate that our approach is able to get significant improvement in performance over leading  MARL approaches  on three benchmark problem domains from literature.
	\vspace{-0.1in}
	\section{Related Work} In this section, we highlight research of relevance to the contributions of this paper. 
	
	\noindent \textbf{Sparse Good Experiences}
	
	\noindent Imitation learning (IL) enables a learner to imitate expert behavior in an underlying MDP environment. A wide variety of IL methods have been proposed in the last few decades. The simplest IL method among those is Behavioral Cloning (BC)~\cite{pomerleau1991efficient} which: (i) collects demonstrations from expert(s); (ii) treat the demonstrations as i.i.d state-action pairs; (iii) learn policy using supervised learning.  ~\cite{lerer2019learning} is another BC approach that is focussed on social dilemma. BC requires many demonstrations and unfortunately, it is  typically not feasible to obtain many demonstrations from experts in real-world scenarios.
	
	\cite{lee2019improved} employs demonstrations to improve multiagent learning. This paper is limited to problem settings where reasonable centralized policy can be obtained, therefore their method is only applicable to small 2 agents problems for which they can compute a centralized policy either from MMDP or MPOMDP. Since problems considered in the paper have more number of agents, it is not feasible to solve an MMDP or MPOMDP to obtain a centralized policy.
	
	~\cite{oh2018self} provides a Self Imitation Learning (SIL) approach for single agent RL where (good) experiences generated during exploration are stored in a prioritized buffer (henceforth referred to as the $M_{SI}$ buffer) based on cumulative reward achieved. During training, it samples the experiences from this buffer and trains the neural networks only if the network is predicting a lower value for these experiences. SIL does not directly extend to multi-agent RL and in this paper, we provide an extension of SIL for multi-agent settings. 
	
	%\cite{lerer2019learning} is one of the work that is focused on social dilemma but not explicitly on handling sparse rewards. It is a behaviour cloning approach.
	%Behavior cloning approaches of the kind suggested by the reviewer require presence of expert data and idea is to train agents to mimic them. Reviewer should note that we focus on self imitation (where expert data is not required) and not on ``imitation of expert''.
	%Since self-imitation learns to mimic good experiences all by itself, it does not require expert data and is more scalable (w.r.t number of agents)

	\noindent \textbf{Non-Stationarity}
	
	\noindent There are two threads of relevant research in cooperative MARL for dealing with non-stationarity. First, we have team learning approaches~\cite{haynes1995evolving,claus1998dynamics} where a single learner learns policies for a team of agents.  Team learning approaches suffer from curse of dimensionality. Furthermore, it may not be realistic to assume centralization of information, especially if the agents themselves receive decentralized observations that cannot be shared with other agents at every step. 
	
	The second thread of research has focussed on decentralized learning~\cite{agogino2006quicr,tampuu2017multiagent}, where agents learn concurrently to avoid the curse of dimensionality and centralization of information. Since individual agents are changing their policies concurrently, RL problem experienced by each agent is no longer stationary and can result in unstable and divergent learning performance.   In order to address this non-stationarity issue, a centralized critic is employed. One of the leading approaches in this space is called COMA~\cite{foerster2018counterfactual}.  
	
	~\cite{palmer2018lenient} have applied ``leniency" and \cite{omidshafiei2017deep} have applied Hysteric Q Learning to counter non-stationarity problem in MARL.  Unfortunately, none of these approaches have a mechanism for handling the issues of exploration and forgetting of good policies arising due to having only a few good experiences. 
	
	\noindent \textbf{Scalability}
	
	\noindent A leading approach is by ~\cite{nguyen2017policy} to solve cooperative problems with large numbers of homogeneous agents and anonymous interactions. However, it relies on having non-global states and transition function decomposability given number of agents. This is not feasible in domains of interest in this paper and since it is based on actor critic architecture, it has same issues as other MARL approaches with sparse rewards. 
	
	Another approach (~\cite{yang2018deep}) along this line is based on mean field games~\cite{lasry2007mean}. Unfortunately, approaches based in mean field, where indistinguishably property should hold - i.e, the game should be invariant under permutation of the agents' indices, are not suitable as different types of agents (ambulances and fire trucks) can exist in MARL problems. 
	
	The last thread of relevant research has employed game theory to develop decentralized learning methods~\cite{hu2003nash,heinrich2015fictitious,heinrich2016deep}. One of the leading approaches is the neural fictitious self play method~\cite{heinrich2016deep}, which employs ideas from the well known fictitious play~\cite{Brownetal} method. Given the focus on equilibrium for game theoretic methods, these approaches can get stuck in bad local optima in the case of cooperative problems. However, a key advantage of relevance  specifically of NFSP is being able to perform decentralized learning at scale.

	\section{Background}
	
	In this section, we describe key concepts/approaches on which we build upon in this paper, namely Generalized Weakened Fictitious Play and Neural Fictitious Self Play (NFSP). 
	
	\subsection{Generalized Weakened Fictitious Play, GWFP}
	\label{sec:gwfp}
	In Fictitious play (FP), a popular approach for computing Nash Equilibrium in normal-form single shot games, fictitious players choose exact best responses against  their opponents' average strategy at each iteration. FP is guaranteed to converge to a Nash equilibrium for zero-sum games, potential games and identical interest games (i.e., cooperative multi-agent problems).  FP requires the computation of exact best response and \cite{leslie2006generalised}  relaxed this requirement by providing Generalized Weakened Fictitious Play (GWFP). GWFP works with approximate best responses as follows:
	$$\pi^{t+1}  \in (1 - \eta^{t+1}) \pi^t + \eta^{t+1} \cdot b_{\epsilon^t}(Q^t)$$
	where $\eta^t \rightarrow 0$, $\epsilon^t \rightarrow 0$, $||Q^{t} - R(\pi^t)|| \rightarrow 0$ as $t \rightarrow \infty$.
	$b_{\epsilon^t}(Q^t)$ is best response to the policy. $R(\pi^t)$ is the reward for an agent given that it is following policy $(\pi^t)$.
	This generalized and weakened version has similar guarantees as the original FP algorithm and converges for potential games, identical interest and zero sum games.  
	
	%FP and its weakened definition, GWFP above are typically associated with a normal-form representation, where each agent acts only once per one game, which is not suited to real problems where each agent takes multiple decisions and there are multiple such agents. NFSP (described below) was developed to address this issue. 
	
	\subsection{Neural Fictitious Self Play (NFSP)}
	In order to overcome the scalability issue (particularly with respect to agents taking multiple decisions) with FP and its extensions.~\cite{heinrich2015fictitious,heinrich2016deep} proposed an appropriately approximated method for generalized weakened fictitious play referred to as Neural Fictitious Self Play (NFSP). 
	Specifically, 
	\squishlist 
	\item Instead of computing the exact best response strategy, NFSP learns an approximate best response using Deep Q-Networks (DQN)~\cite{mnih2015human}. Deep Q network with parameters $\theta^Q$ is trained using the following loss function: 
	{\small $$\resizebox{.98\hsize}{!}{$ {\cal L}(\theta^Q) = \mathbb{E}_{(s,a,r,s')\sim {\cal M}_{RL}} \Big[ \Big(r + \max_{a'} Q(s',a' | \theta^{Q'}) - Q(s,a | \theta^Q)\Big)^2\Big]$} $$ }
	where ${\cal M}_{RL}$ refers to the stored RL experiences (i.e., past game transitions).
	\item Instead of averaging full exact strategies, each agent learns an approximate average strategy by using supervised learning (SL) with deep neural networks~\cite{heinrich2016deep} :
	{\small $${\cal L}(\theta^{\Pi}) = \mathbb{E}_{(s,a)\sim {\cal M}_{SL}} \Big[ - log(\pi(s,a | \theta^{\Pi}) \Big]$$}\\
	where ${\cal M}_{SL}$ refers to the stored Supervised Learning experiences (i.e., past best responses).
	\squishend

	\section{Neural Fictitious Self Imitation Play, NFSIP}
	%\begin{figure*}
	%%	\includegraphics[width=2in, height=3in]{images/Flow_1.JPG}
	%%	\includegraphics[width=3.5in, height=3in]{images/Flow.JPG}
	%%	\includegraphics[width=5.5in, height=3in]{images/Flow_2.JPG}
	%	\includegraphics[width=6.8in, height=2.5in]{images/Flow_4.JPG}
	%	\vspace{-0.1in}
	%	\caption{NFSIP}
	%	\label{fig:Flow_NFSIP}
	%\end{figure*}
	
	In this section, we describe our main algorithm, NFSIP (pseudocode in Algorithm~\ref{alg:NFSIP}) for cooperative MARL problems in the presence of only a few good experiences. Here are the key contributions in NFSIP:\\
	\noindent 1. NFSIP implements self imitation for multi-agent settings in context of NFSP. \\
	\noindent 2. NFSIP provides novel insights on value and policy network updates that preserve the fictitious play property while dealing with the issue of few good experiences. This ensures guarantees on convergence under certain conditions.  \\
	
	\noindent There are four key steps to the NFSIP algorithm:\\
	\noindent 1. \textbf{Store experiences in appropriate replay buffers}:An NFSIP agent interacts with its fellow agents and stores its experience of state transitions in $M_{RL}$ buffer and its own best response behaviour in $M_{SL}$ buffer (lines 6 in pseudocode). Once an episode ends, a copy of the individual experiences, $(s,a,r,s')$ updated to include cumulative rewards R (i.e., $(s,a,R(s,a),s')$) are stored in the prioritized buffer, $M_{SI}$ . 
	Once episode ends, in lines 10-14, we update the self imitation buffer, $M_{SI}$ with experiences if social welfare (welfare of the entire system, including all agents) is higher than the set threshold for social welfare (bestReward achieved so far). These experiences are updated to include cumulative rewards. \\
	\noindent 2. \textbf{Learn from all experiences}: For each agent, in NFSIP, we update  the average policy network and Q-network parameters based on all the experiences (good and bad).  \\
	\vspace{-0.1in}
	{\small $${\cal L}(\theta^Q) = \mathbb{E}_{(s,a,r,s')\sim {\cal M}_{RL}} \Big[ \Big(r + max_{a'} Q(s',a' | \theta^{Q'}) - Q(s,a | \theta^Q)\Big)^2\Big] $$}
	{\small ${\cal L}(\theta^{\Pi}) = \mathbb{E}_{(s,a)\sim {\cal M}_{SL}} \Big[ - log(\pi(s,a | \theta^{\Pi}) \Big]$} \\
	Q and $\Pi$ networks are updated in lines 8-9 of pseudocode. \\
	\noindent 3. \textbf{Learn from self imitation buffer}: Since there are only a few good experiences, it is imperative that updates from ``good" experiences (i.e., ones that improve social welfare) are not overwritten by ``bad" experiences. Therefore, in NFSIP, we have a separate self imitation loop at the end of each episode to not forget the learning from ``good" experiences. In this self imitation loop, we update both the average policy and Q-network parameters based on difference in the reward obtained from the episode and the current value function estimate. 
	{\small 
		$${\cal L}(\theta^Q) = \mathbb{E}_{(s,a,R,s')} \Big[([R(s,a) - V(s|\theta^Q)]_+)^2\Big]$$}
	{\small $${\cal L}(\theta^{\pi}) =  \mathbb{E}_{(s,a,R)} \Big[-log(\pi(s,a|\theta^\Pi)) \cdot [R(s,a) - V(s|\theta^Q)]_+)\Big] $$}
	where, {\small$[R(s,a) - V(s|\theta^Q)]_+ = \max(0, R(s,a) - V(s|\theta^Q))$} \\
	In lines 18-19 of pseudocode, $Q$ and $\Pi$ networks are trained with experiences from self imitation buffer, $M_{SI}$ if $Q$-network is predicting a lower value for these experiences as compared to their actual (cumulative) reward.\\
	\noindent 4. \textbf{Mixing the approximate average strategy and approximate best response}:  The resulting Q-network (from the above parameter updates) for each agent is used in its approximate best response strategy, $b_{\epsilon}(Q)$, which selects a random action with probability $\epsilon$ and otherwise chooses the action that maximizes the predicted action values. On the other hand, we have the $\Pi$-network  which defines the agents' average strategy so far. During execution, the agent chooses its actions from a mixture of its two strategies, $b_{\epsilon}(Q)$ and $\Pi$.
	Line 4 ensures the mixing of average and approximate best response using the parameter $\eta$. \\
	Having only a few good experiences results in bad multi-agent cooperative learning for two reasons: (i) Good experiences are few so value updates can be lost due to bad experiences and non-stationarity. (ii) Policy averaging can result in bad policies overwriting the impact of good policies. To that end, we provide two sets of novel insights in NFSIP with respect to steps 3 and 4 above that help in learning good policies and good value functions even when there are only a few good experiences: 
	\squishlist
	\item Self Imitation Learning for Cooperative MARL: This is to ensure value updates corresponding to good experiences happen multiple times if the value is learned incorrectly for states involved in good experiences. 
	\item Good experience driven policy averaging: This provides a novel way of weighted policy averaging in Fictitious Play to ensure good policies are not washed away. 
	\squishend
	
	\subsection{SIL for Cooperative MARL} 
	%As indicated in step 3 of NFSIP, we employ Self-Imitation buffer to ensure good experience related parameter updates are not overwritten. 
	Self imitation learning in single agent case imitates past good experiences multiple times (based on priority) and prioritizes learning with those good experiences. However, in multi-agent problems, due to simultaneous learning of agents, past good experience for an agent may not be a good experience if other agents have changed their policy. Therefore, our \textbf{\em first} insight here is to judge the goodness of any experience not just based on its own reward but also based on social welfare.
	
	Due to non stationary environment we want to avoid utilizing old experiences, for this we periodically remove expert data (self) generated for self imitation process. We do so when we encounter a better social welfare solution. Our \textbf{\em second} insight here is to employ a threshold value that is slowly  adjusted to ensure that there are always expert experiences for training that are not too old and provide higher social welfare. 
	
	Finally, we train only with experiences where neural network is predicting a lower value than the actual value (cumulative reward) of the agent. To ensure this, we employ the following term in value and policy parameter updates: 
	$$\resizebox{.98\hsize}{!}{$ [R(s,a) - V(s|\theta^Q)]_+ = 
		\begin{cases}
		max(0, (R(s,a) - V(s|\theta^Q))) &\text{ if } W >= W_T \\
		0  &\text{otherwise}
		\end{cases} $} $$
	Where $R(s,a)$ = \text{Cumulative reward of agent},  \\ 
	$W = \sum R(s,a)$ i.e, \text{Welfare of the entire system (social welfare)} \\ 
	$W_T$ = \text{Threshold value for social welfare} 
	\subsection{Good Experience Driven Policy Averaging}
	
	We first highlight the key issue with policy mixing in NFSP with regards to sparse rewards.  NFSP employs maximum log likelihood (using loss as negative log likelihood) for learning the mixture of past policy, $\pi^t$ and current approximate best response policy,  $b_{\epsilon^t}(Q^t)$ based on the observed samples (i.e., best response actions taken at each iteration).  The standard maximum likelihood principle implicitly places equal weight on each of the observations in the sample.  Taking the example of coin toss, if after 1000 iterations, if we observed 700 heads and 300 tails, maximum likelihood will predict a biased coin with 0.7 and 0.3 probability. However, this is incorrect as the sampled data was biased. This issue is more prominent in RL problems where good experiences come by rarely.  So, samples data is bound to have rare occurrences of them, causing maximum likelihood to result in bad local optima. 
	
	One way to improve the model is to use weighted maximum likelihood. Such methods have been employed in for risk management in Finance ~\cite{steude2011weighted} and for image denoising in image processing~\cite{deledalle2009iterative}).~\cite{steude2011weighted} have shown that  downweighting the observations that bear a high probability of being destructive outliers can considerably improve the forecast accuracy for a variety of data sets and different time series models. \cite{deledalle2009iterative} derived the weights in a data driven manner. The weights are iteratively refined.
	
	For solving MARL with a few good experiences, we build on similar ideas. Specifically,  we increase weight for better experiences. These weights are dynamically updated based on the current state of learning. Since we only want to increase the weight of good experiences, we will not have negative weights. For average policy network, we employ the following additional loss based on experiences in $M_{SI}$:\\
	$\mathbb{E}_{(s,a,R)\sim M_{SI}} \Big[-log(\pi(s,a)) \cdot [R(s,a) - V(s)]_+\Big]$ \\
	$\text{where,} \quad [R(s,a) - V(s)]_+ = max(0, R(s,a) - V(s))$ \\
	On similar lines, we also add an additional weight to the Q-network loss based on self imitation memory, $M_{SI}.$\\
	
	\noindent \textbf{Theoretical Intuition}\\
	\noindent In this section, we provide the intuition for why good experience driven policy averaging in NFSIP satisfies the GWFP property of Section~\ref{sec:gwfp}.  This is an important property as it justifies the convergence of NFSIP for cooperative MARL problems. 
	
	\noindent Specifically, we show that if policy averaging in NFSP is:
	{\small $$\pi^{t+1} \in (1 - \eta^{t+1}) \pi^t + \eta^{t+1}\cdot b_{\epsilon^t}(Q^t), \textbf{with } \eta^{t+1} = \frac{1}{t+1}$$
		then, policy averaging in NFSIP is given by
		$$\pi^{t+1} \in (1 - \eta_{NFSIP}^{t+1}) \pi^t + \eta_{NFSIP}^{t+1}\cdot b_{\epsilon^t}(Q^t), \textbf{with } \eta_{NFSIP}^{t+1} = \frac{1 + \Gamma}{t+1}$$
		$$\textbf{ and } \Gamma = [R(s,a) - V(s | \theta^Q)]_{+}$$}
	\noindent \textbf{\em Intuitively, this is to say that NFSIP just changes the mixing parameter (that satisfies all properties desired of the mixing parameter) in comparison to NFSP. }\\
	
	\noindent \underline{\textbf{\em Policy Averaging in NFSP:}}
	We begin with NFSP network updates for policy averaging in NFSP.  The action-value network loss function is given by: 
	{ \small \begin{align}
		{\cal L}(\theta^Q) = \mathbb{E}_{(s,a,r,s')} \Big[\Big( r(s,a) + max_{a'} Q(s',a' | \theta^{Q'}) - Q(s,a | \theta^Q)\Big)^2\Big] \label{FP_Q_Loss}
		\end{align}}
	\noindent The policy network loss function is given by:
	{ \small \begin{align}
		{\cal L}(\theta^{\pi}) = \mathbb{E}_{(s,a)} \Big[ - log(\pi(s,a | \theta^{\Pi}) \Big] \label{FP_P_Loss}
		\end{align}}
	When we train the two networks (learning rates $\alpha$,$\beta$), the parameter updates for policy and action-value networks are as follows: \\
	\noindent {\textbf{\em $\Pi$-network update}:} 
	\vspace{-0.25in}
	{ \small \begin{align}
		\hspace{0.5in}\theta^{\Pi} &= \theta^{\Pi} + \alpha  \nabla log (\pi(s,a | \theta^{\Pi}) \label{FP_P_Update}
		\end{align}}
	\noindent {\textbf{\em Q-network update}:}
	{ \small \begin{align}
		&\theta^{Q} = \theta^Q - \beta \cdot \nabla \Big( r + max_{a'} Q(s',a' | \theta^{Q'}) - Q(s,a | \theta^Q)\Big)^2 \nonumber \\
		\intertext{Since $Q(s',a' | \theta^{Q'})$ is based on $\theta^{Q'}$ and not $\theta^Q$}
		%&= \theta^{Q} - \beta . 2\Big( r + max_{a'} Q(s',a' | \theta^{Q'}) - Q(s,a | \theta^Q)\Big) \nabla (- Q(s,a | \theta^Q)) \nonumber \\
		&= \theta^{Q} + \beta . 2\Big( r + max_{a'} Q(s',a' | \theta^{Q'}) - Q(s,a | \theta^Q)\Big) \nabla Q(s,a | \theta^Q) \label{FP_Q_Update}
		\end{align}}
	\noindent GWFP~\cite{leslie2006generalised} is defined as follows:
	$$\pi^{t+1}  \in (1 - \eta^{t+1}) \pi^t + \eta^{t+1} \cdot b_{\epsilon^t}(Q^t)$$
	where $\eta^t \rightarrow 0$, $\epsilon^t \rightarrow 0$, $||Q^{t} - R(\pi^t)|| \rightarrow 0$ as $t \rightarrow \infty$
	
	\noindent NFSP and standard FP typically employ:  $\eta^{t+1} = \frac{1}{(t+1)}$ in order to satisfy GWFP criterion above.  \\
	
	\noindent \underline{\textbf{\em Network updates with only Self Imitation Learning (SIL):}}
	
	NFSIP employs self imitation loop on top of NFSP updates. We first compute the self imitation learning related updates and add it over the updates above for NFSP. 
	
	\noindent The action-value network loss function is given by: 
	{\small \begin{align}
		{\cal L}(\theta^Q) = \mathbb{E}_{(s,a,R,s')} \Big[([R(s,a) - V(s|\theta^Q)]_+)^2\Big]  \label{SIL_Q_Loss}
		\end{align}}
	
	\noindent The policy network loss function is given by:
	{\small \begin{align}
		{\cal L}(\theta^{\pi})= \mathbb{E}_{(s,a,R)} \Big[-log(\pi(s,a|\theta^\Pi)) \cdot [R(s,a) - V(s|\theta^Q)]_+)\Big] \label{SIL_P_Loss}
		\end{align}}
	Where, $[R(s,a) - V(s|\theta^Q)]_+ =$ \\ $$\begin{cases}
	0,  & \text{if} \quad [R(s,a) - V(s|\theta^Q)] \le 0  \\
	[R(s,a) - V(s|\theta^Q)], &otherwise
	\end{cases}$$
	
	The parameter updates for the two networks are as follows:\\
	%In training of policy network:
	%\begin{itemize}
	%	\item We want to up-weight the good experiences by some positive weight.
	%	\item Which is $[R(s,a) - B(s|\theta^Q)]_+$ in this case.\
	%	\item In every training step  $[R(s,a) - B(s|\theta^Q)]_+$ could have different values based on the current state of network and how baseline is computed. Assuming, $B(s|\theta^Q) =  (1/|A|) \sum_a Q(s,a | \theta^Q)$
	%\end{itemize}
	
	\noindent {\textbf{\em $\Pi$-network update}:}
	
	For a baseline, $V(s | \theta^Q)$ that is independent of current policy, $[R(s,a) - V(s|\theta^Q)]_+$ is constant. Therefore, {\small \begin{align}
		\theta^{\Pi} &= \theta^{\Pi} + \alpha ([R(s,a) - V(s|\theta^Q)]_+) \nabla log (\pi(s,a | \theta^{\Pi}) \label{SIL_P_update}
		\end{align}}
	
	\noindent {\textbf{\em Q-network update}:}
	
	\noindent Q network is optimized based on $[R(s,a) - V(s|\theta^Q)]_+$ :\\
	\textbf{Case 1}: $[R(s,a) - V(s|\theta^Q)]_+ = 0$: This is trivial as there will be no update to Q network. \\
	\textbf{Case 2}: $[R(s,a) - V(s|\theta^Q)] > 0 \implies [R(s,a) - V(s|\theta^Q)]_+ = [R(s,a) - V(s|\theta^Q)]$\\
	
	\noindent Considering $V(s|\theta^Q) = (1/|A|) \sum_a Q(s,a | \theta^Q)$, we have  
	{ \small \begin{align}
		& \nabla ([R(s,a) - V(s|\theta^Q)]_+)^2 = \nabla ([R(s,a) - (1/|A|) \sum_a Q(s,a | \theta^Q)])^2  \nonumber \\
		%&= 2  ([R(s,a) - (1/|A|) \sum_a Q(s,a)]) \nabla ([R(s,a) - (1/|A|) \sum_a Q(s,a | \theta^Q)]) \nonumber \\
		&= 2  (1/|A|) ([R(s,a) - (1/|A|) \sum_a Q(s,a | \theta^Q)]) \nabla (-Q(s,a | \theta^Q)) \nonumber \\
		%&= 2  (1/|A|)  ([R(s,a) - (1/|A|) \sum_a Q(s,a | \theta^Q)]) \nabla (-Q(s,a | \theta^Q)) \nonumber \\ 
		&= 2  (1/|A|)  ([R(s,a) - V(s| \theta^Q)]) \nabla (-Q(s,a | \theta^Q))  
		\end{align}}
	Therefore,
	{ \small \begin{align}
		\theta^{Q} &= \theta^{Q} + \beta . 2 (1/|A|)  ([R(s,a) - V(s|\theta^Q)]_+) \nabla Q(s,a | \theta^Q) \label{SIL_Q_update}
		\end{align}}
	
	\begin{figure*}[htbp]
		\vspace{-0.2in}
		\includegraphics[width=2.5in,height=2in]{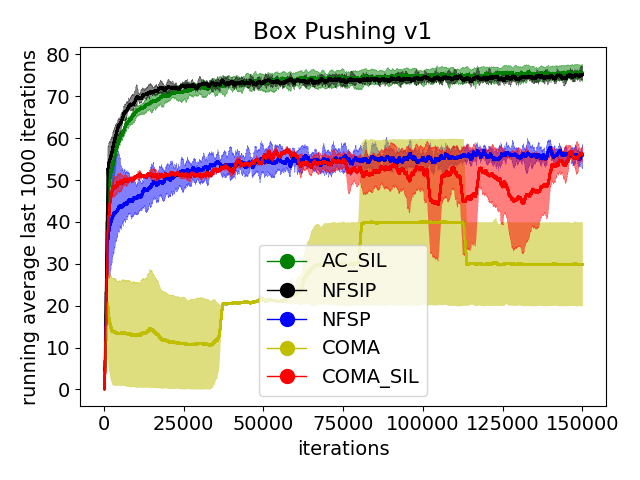}
		\includegraphics[width=2.5in,height=2in]{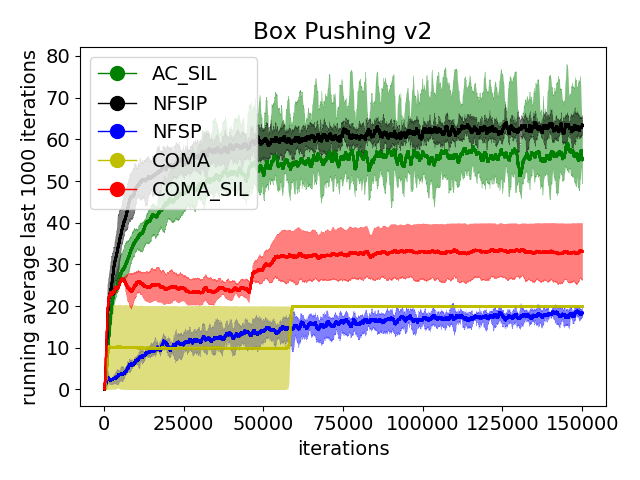}
		\includegraphics[width=2.5in,height=2in]{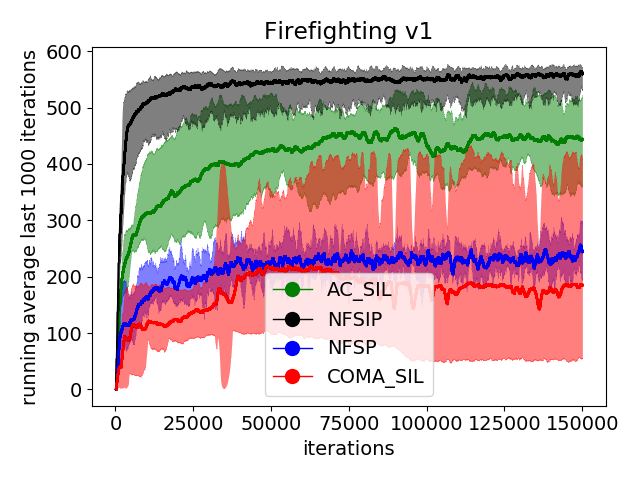}
		\vspace{-0.15in}
		\includegraphics[width=2.5in,height=2in]{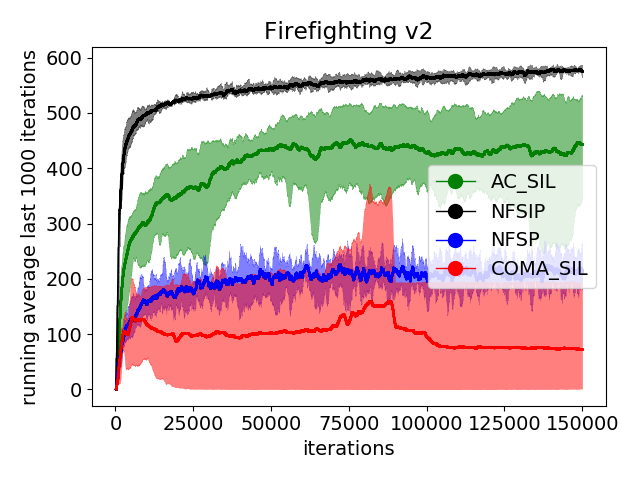}
		\includegraphics[width=2.5in,height=2in]{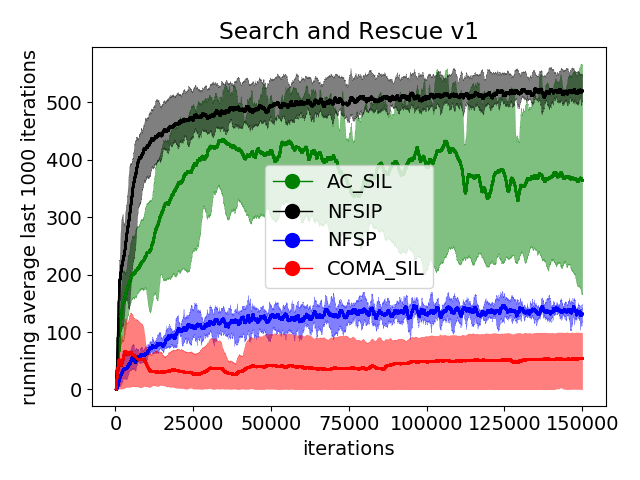}
		\includegraphics[width=2.5in,height=2in]{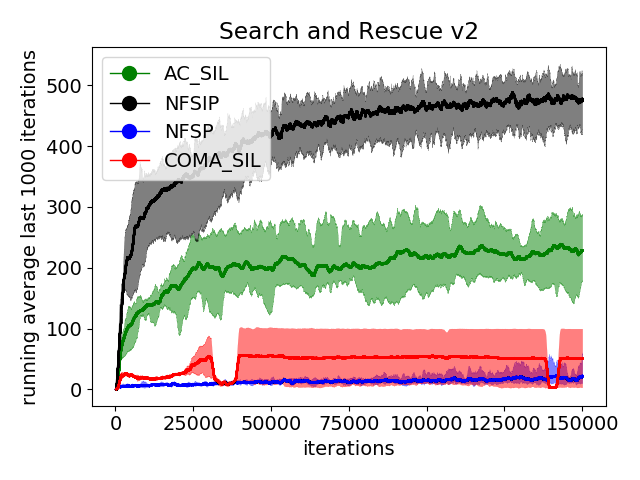}
		\caption{Grid Size: 4x4, Box pushing v1 and v2, Fire Fighting v1 and v2, Search and Rescue v1 and v2: Shaded region represents variance when run the same experiments multiple times. Y axis represents running average of social welfare.}
		\vspace{-0.1in}
		\label{fig:Results4x4}
	\end{figure*}
	
	\noindent \underline{\textbf{\em Policy Averaging for NFSIP:}}
	
	\noindent We now combine NFSP (\ref{FP_P_Update},~\ref{FP_Q_Update}) and SIL updates (\ref{SIL_P_update},~\ref{SIL_Q_update}).\\
	
	\noindent {\textbf{\em $\Pi$-network update}:}
	{\small \begin{align}
		&\theta^{\Pi} = \Big( \theta^{\Pi} - \alpha  \nabla (-log (\pi(s,a | \theta^{\Pi}))\Big) \nonumber \\ 
		&\hspace{0.4in}- \alpha ([R(s,a) - V(s|\theta^Q)]_+) \nabla (-log (\pi(s,a | \theta^{\Pi})) \nonumber \\
		%&= \Big( \theta^{\Pi} + \alpha  \nabla log (\pi(s,a | \theta^{\Pi}) \Big) + \alpha ([R(s,a) - V(s|\theta^Q)]_+) \nabla log (\pi(s,a | \theta^{\Pi}) \nonumber \\
		&=  \theta^{\Pi} + \alpha  (1+ [R(s,a) - V(s|\theta^Q)]_+) \nabla log (\pi(s,a | \theta^{\Pi})  \label{NFSIP_P_update}  
		\end{align}}
	
	\noindent {\textbf{\em Q-network update}:}
	{\small \begin{align}
		\theta^{Q} &= \Big(\theta^{Q} + \beta . 2( r + max_{a'} Q(s',a' | \theta^{Q'}) - Q(s,a | \theta^Q)) \nabla Q(s,a | \theta^Q) \Big) \nonumber \\
		& \hspace{0.1in}+ \beta . 2 (1/|A|) \cdot ([R(s,a) - V(s|\theta^Q)]_+) \nabla Q(s,a | \theta^Q) \nonumber \\
		&= \theta^{Q} + \beta . 2\Big( r + max_{a'} Q(s',a' | \theta^{Q'}) - Q(s,a | \theta^Q(s,a | \theta^Q)) \nonumber \\ 
		& \hspace{0.1in} + (1/|A|) \cdot [R(s,a) - V(s|\theta^Q)]_+ \Big) \nabla Q(s,a | \theta^Q) \label{NFSIP_Q_update}
		\end{align}}
	
	\noindent NFSP satisfies GWFP with $\eta = \frac{1}{t+1}$. With policy and Q update as given in ~\ref{NFSIP_P_update} and ~\ref{NFSIP_Q_update} respectively, NFSIP satisfies GWFP property in the same way as NFSP:
	$$\pi^{t+1}  \in (1 - \eta_{NFSIP}^{t+1}) \pi^t + \eta_{NFSIP}^{t+1} \cdot b_{\epsilon^t}(Q^t)$$
	with $\eta^{t+1}_{NFSIP} = ((1 + \Gamma)/{(t+1)})$ and $\Gamma = [R(s,a) -V(s|\theta^Q)]_+$,\\
	where $\eta_{NFSIP}^{t+1} \rightarrow 0$, $\epsilon^{t+1} \rightarrow 0$, $||Q^{t+1} - R(\pi^{t+1})|| \rightarrow 0$ as $t + 1 \rightarrow \infty$
	
	GWFP holds exactly when the baseline, $V(s | \theta^Q)$ is independent of policy. However, when the baseline is dependent on policy (e.g., $V(s | \theta^Q) = \sum_{a} \pi(s,a | \theta^{\Pi}) Q(s,a | \theta^Q)$ ), there is an additional term with respect to policy, $\pi$ in the update expression of Q-network. In practice, we see that the performance converges in all our examples when we use a baseline dependent on policy. 
	
	\begin{figure}[!h]
		\includegraphics[width=1.65in,height=1.4in]{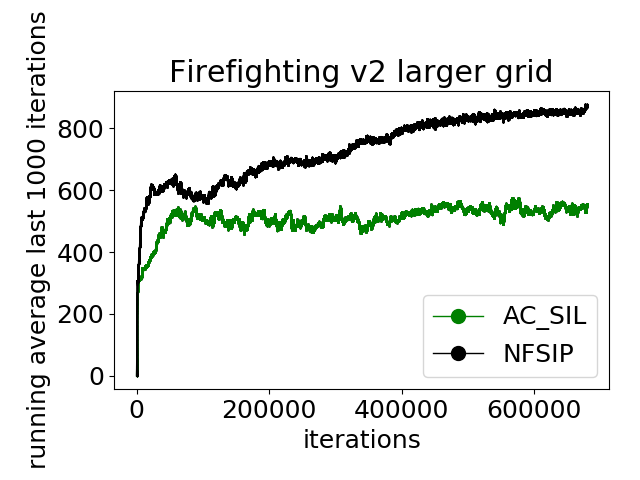}
		\includegraphics[width=1.65in,height=1.4in]{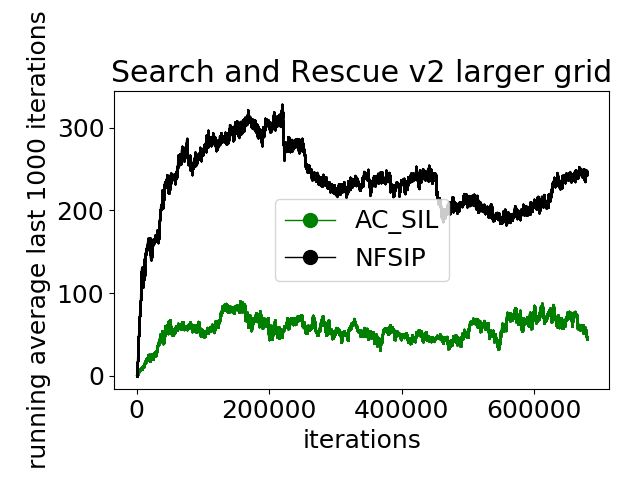}
		\vspace{-0.15in}
		\caption{Grid Size = 6x6, Fire Fighting v2, Search and Rescue v2. Y axis represents running average of social welfare}
		\label{fig:Results6x6}
	\end{figure}

	{\small \begin{algorithm}[!h]
			\caption{$\text{Neural Fictitious Self Imitation and Play, NFSIP}$}
			\label{alg:NFSIP}
			\begin{algorithmic}[1]
				\STATE Initialize $\theta^\Pi$, $\theta^Q$ and $\theta^{Q'}$ networks \\
				\STATE bestReward = $-\infty$ \\
				\WHILE {Not Converged}
				\STATE policy =
				$\begin{cases}
				b_{\epsilon}(Q)  & \text{with probability} \quad\eta \\
				\pi  & \text{with probability}\quad 1- \eta
				\end{cases}$ 
				\FOR {every time step}
				\STATE Simulate agents and store experiences in $M_{RL}$ and $M_{SL}$ (if agent took best response action)
				\FOR {all agents} \STATE Sample from $M_{RL}$, train $\theta^Q$ using Q-Loss (Eq \ref{FP_Q_Loss})\STATE Sample from $M_{SL}$, train $\theta^\pi$ using $\pi$-Loss (Eq~\ref{FP_P_Loss})
				\ENDFOR
				\ENDFOR
				
				\IF {episodeReward $>$ bestReward}
				\STATE Reset $M_{SI}$ and bestReward = Episode reward
				\ENDIF
				\IF {episodeReward $>=$ bestReward}
				\STATE Compute cumulative reward, $R$
				\STATE Store experiences in $M_{SI}$ prioritized on $R$ 
				\ENDIF	
				\FOR {some iteration}  
				
				\FOR {all agents}
				
				\STATE Sample from prioritized replay buffer, $M_{SI}$
				\STATE Train $\theta^Q$ using SIL Q-loss (Eq~\ref{SIL_Q_Loss})
				\STATE Train $\theta^\pi$ using SIL $\pi$-loss (Eq~\ref{SIL_P_Loss})
				%\STATE $[R(s,a) - V(s)]_+ = \max \big(0, R(s,a) - V(s)\big)$
				%\STATE $V(s) = \sum_a \pi(s,a) \cdot Q(s,a)$
				\ENDFOR
				\ENDFOR
				
				\STATE Update target action-value network, $\theta^{Q'}$ periodically
				\ENDWHILE
			\end{algorithmic}
	\end{algorithm}}
\vspace{-0.1in}
\section{Experimental Results}

In this section, we evaluate the performance of our approach (NFSIP) in comparison to leading approaches for cooperative MARL. We perform the comparison on three different benchmark problems from literature: (a) Box Pushing~\cite{seuken2012improved}; (b) Fire Fighting~\cite{oliehoek2008optimal}; and, (c) Search and Rescue~\cite{nanjanath2010decision,parker2016exploiting}. We extend these problem settings to ones with many agents and larger state space, so as to make good experiences sparse.  We compare against the following leading approaches for cooperative MARL: (a) COMA; (b) NFSP; (c) AC-SIL: Multi-agent extension of SIL; (d) COMA\_SIL: An SIL extension for COMA. 

We now provide details of the benchmark problems:\\
\noindent -\textbf{Box pushing problem}~\cite{seuken2012improved}: Multiple agents need to coordinate and push boxes of different sizes to their goal locations in a grid world. Each agent has 6 possible actions to take: \{move left, move right, move up, move down, act on the task, stay\}. To successfully push a box, certain number of agents need to act on it. For this domain, we created simpler instances with a 4x4 grid, 4 boxes and 5-agents in box pushing.  We created different versions of this problem\footnote{(V1) Any single agent can push the box; and (V2): To push any box at least 2 agents need to cooperate and simultaneously act on it.} with smaller grid sizes as benchmark algorithms were unable to learn at all on larger problem instances. \\
\noindent -\textbf{Firefighting problem}~\cite{oliehoek2008optimal}: In this problem setting we have a 4x4 grid with 10  agents (fire trucks), fires are spread over different locations. Number of trucks needed to put out the fire depends on its intensity (low/high). We created different versions of the problem\footnote{ (V1): 2 agents can put out the fire with probability 0.9, more than 2 agents can do so with probability 1; and (V2): Intensity of fire will increase from low to high with probability 0.2 at every time step. Low intensity fire: ``2 agents can put it out probability 0.9, more than 2 agents with probability 1". High intensity fires: ``2 agents can put it out with probability 0.75, 3 agents can put it out with probability 0.9 and more than 3 agents can do it with probability 1".}\\
\noindent -\textbf{Search and Rescue}~\cite{parker2016exploiting}: Different types of agents (such as firetrucks and ambulances) need to coordinate with each other. In this problem setting we have a 4x4 grids with 5 ambulances and 5 firetrucks. Number of firetrucks and ambulances needed to complete the task depends on difficulty of the scenario. We created different versions of the problem\footnote{(V1): Minimum 1 fire truck and 1 ambulance needs to cooperate to complete the task; and (V2): Difficulty of the search and rescue scenario will increase from low to high with probability 0.2 if operation is not completed. If difficulty level is low then minimum 1 ambulance and 1 fire truck can complete search and rescue, if difficulty level is high then minimum 2 ambulances and 2 firetrucks are needed to carry out the operation. }

%Before we provide the main results, we show that the counterfactual way of handling the credit assignment fares badly in problems of interest in this paper. Figure~\ref{fig:credit} shows that COMA without counterfactual performs better on both box pushing problems. We have made the same observation in other problems as well. In COMA without counterfactual version, it should be noted that we used the same credit assignment scheme as we used for our NFSIP approach. 

All results are averaged over multiple runs. We ran NFSIP, NFSP and AC\_SIL for 5 times each. In results we plot average over 5 runs (line plot) as well as variance over different runs (shaded region).  Due to counterfactual baseline computation for every action, COMA is very slow (and took 1-2 weeks for training) as compared to our approach (which took 1-2 days).  Here are the key observations from Figures~\ref{fig:Results4x4} and ~\ref{fig:Results6x6}:
\squishlist
\item On the simplest problems, i.e., ones in box pushing, COMA is able to learn good policies. However, NFSIP and AC\_SIL perform the best even on these simplest problems. 
\item NFSIP is able to outperform both NFSP and COMA on all 6 scenarios
\item NFSIP is able to perform as good as or better than AC\_SIL. In the last scenario (Search and Rescue V2), NFSIP is able to get a result that is 5 times that of AC\_SIL. 
\item NSFIP not only outperformed COMA\_SIL, AC\_SIL and NFSP, variance is also low in case of NFSIP as compared to other approached compared here.
\squishend
\vspace{-0.1in}
\subsection{Neural network Architecture and Training:} Policy/Q network in NFSIP has 2 hidden layers (32 nodes in each layer). We used same number of hidden layers/nodes in all experiments/methods. After every hidden layer we used layer norm. In all experiments we start with exploration rate of 10\% (NFSP/NFSIP: $\eta=0.2$, $\epsilon=0.5$ and $\epsilon=0.2*0.5 = 0.1$ for other methods). After every 500 iteration we reduce epsilon to a factor of 0.98. In NFSP/NFSIP all agents share parameters in both networks. i.e, there is one policy network and one best response network that takes agents Ids as input to distinguish between them. We used learning rate of $10^{-3}$ for actor/policy and $10^{-4}$ Q/Critic network.  We wan SIL loop 5 times (line 19 of Algorithm~\ref{alg:NFSIP}). Hyper parameters were coarsely tuned on the box pushing scenario and then used for firefighting and 'Search and Rescue'. The most sensitive parameter was exploration parameter. In all methods we periodically discarded older experiences (except the SL buffer in NFSP/NFSIP). And used batch training with batch size of 32. 

For tuning the social welfare threshold value we experimented with different techniques, But since here in all problem setting reward is discrete therefore we went with most logical choice of tuning it, which is whenever we encounter the experiences for which social welfare is higher than current threshold, we update the threshold value to current social welfare and discarded old experiences.

\bibliography{Biblography}
\bibliographystyle{named}

\end{document}